\PassOptionsToPackage{sort&compress,numbers}{natbib}
\documentclass{article}

\usepackage[preprint]{corl_2026} 
\usepackage{booktabs}
\usepackage{graphicx}
\usepackage{float}
\usepackage{hyperref}
\usepackage{threeparttable}   
\usepackage{placeins} 

\title{A Durable Vision-Based Tactile Fingertip for Robotic Manipulation}

\author{
  F. Richard Cottrell$^{1}$, Megha H. Tippur$^{1}$, Edward H. Adelson$^{1}$\\
  $^{1}$MIT CSAIL\\
  \texttt{frichardcottrell@gmail.com, <mhtippur, adelson>@csail.mit.edu}
  \vspace{-20pt}
}

\begin{document}
\maketitle


\begin{abstract}
\label{sec:abstract}
Currently available commercial vision-based tactile sensors provide rich contact information but remain vulnerable to abrasion and repeated concentrated loading, limiting their use in demanding robotic applications. This work presents a durable tactile fingertip comprising a soft silicone gel with a nonpigmented, textured, thin thermoplastic-polyurethane protective film and a replaceable sensing cartridge. Durability was evaluated using two accelerated laboratory procedures: a rotating-drum sanding test and a repetitive probe test applying 39.2 N (4.0 kgf) at 45 cycles per minute. Under the defined sanding conditions, the developed sensor reached the protective-film rupture endpoint after approximately 2–3 hours. During repetitive probe testing, all nine developed sensors remained functionally usable when testing was discontinued: seven after 5 days, one after 6 days, and one after 8 days. Commercial GelSight Mini and DIGIT specimens exhibited initial surface-film rupture after approximately 24–30 seconds of sanding and 25–35 minutes of repetitive loading. Damage to the developed sensor progressed gradually and produced little interference with tactile imaging at the test endpoints. These observations establish durability improvements of more than two orders of magnitude under the defined accelerated conditions. Combining increased durability, gradual degradation, and rapid cartridge replacement offers a practical approach to maintainable vision-based tactile sensing for demanding robotic applications.
\end{abstract}



\section{Introduction}
\label{sec:introduction}
	
Tactile sensing provides information that is difficult to obtain from external vision alone, including contact location, local surface geometry, contact-force distribution, shear, and incipient slip. These capabilities are particularly important for robotic hands operating in unstructured environments, where object geometry, position, compliance, and surface condition may not be known in advance \cite{zhang2022hardware, abad2020visuotactile}.
    
Vision-based tactile sensors (VBTSs) use an internal camera to observe deformation of a compliant sensing surface. This approach can provide high spatial resolution and rich contact information using comparatively simple optical and imaging components. GelSight established the use of a camera to image deformation of an illuminated reflective elastomer for high-resolution measurement of contact geometry and force \cite{yuan2017gelsight}. Subsequent designs have produced more compact robotic sensors, including DIGIT \cite{lambeta2020digit}, and soft rounded fingertips suitable for dexterous robotic manipulation \cite{romero2020soft}.
    
Most GelSight-type sensors employ a pigmented or reflective contact surface that suppresses images of the external scene. Other vision–based tactile architectures use transparent or selectively transmissive surfaces to combine tactile sensing with visual or proximity information. FingerVision introduced a transparent marked skin through which internal cameras could observe both contact deformation and the surrounding scene \cite{yamaguchi2019tactile}. See-Through-your-Skin and Finger-STS further developed combined visual, proximity, and tactile perception \cite{hogan2021seeing, hogan2022fingersts}. TIRgel uses total internal reflection to distinguish contact from external visual information \cite{zhang2023tirgel, fan2025crystaltac}. Together, these studies demonstrate several ways in which nonpigmented sensing surfaces can provide both optical contact information and external visual information.

The sensor investigated here also uses a nonpigmented surface and total internal reflection, but with a finely textured outer surface that emphasizes objects in contact with, or immediately adjacent to, the sensing surface while reducing distracting images of more distant objects. The rounded fingertip geometry was influenced by previous soft, rounded GelSight fingertip designs \cite{romero2020soft}. The present work focuses not primarily on introducing another tactile-imaging principle, but on improving the durability and maintainability of the replaceable sensing surface.

Mechanical durability remains an important challenge for VBTS deployment. The compliant elastomeric surface must deform sufficiently to preserve tactile sensitivity while resisting abrasion, repeated concentrated loading, cutting, tearing, and interfacial delamination \cite{abad2020visuotactile, zhang2023tirgel}. Increasing hardness or protective-layer thickness can improve resistance to damage but may reduce compliance or degrade tactile-image quality, creating a trade-off between sensitivity and service life.

Several recent studies have addressed aspects of this problem. Rayamane et al. \cite{rayamane2022design} added a latex protective layer to improve the robustness of a silicone-based optical tactile sensor. PolyTouch employed a protected, replaceable tactile structure and reported substantially longer lifetime than commercial tactile sensors under repeated rubbing \cite{zhao2025polytouch}. Davis and Stuart compared silicone and polyurethane VBTS materials using repeated compression, shear, and abrasion protocols and demonstrated the potential of polyurethane for improving surface resilience \cite{davis2025benchmarking}. These studies establish growing interest in tactile-surface durability, but the literature still contains limited information on extended repetitive concentrated loading, progressive failure behavior, and maintenance-oriented sensor construction. Directly comparable field-lifetime data from industrial robotic operation also remain limited.

For industrial applications, durability alone is insufficient. A sensing surface will eventually wear, and its practical value therefore also depends on how damage develops and how easily the worn component can be replaced. Gradual degradation that preserves useful tactile information can permit maintenance to be scheduled before complete loss of sensing capability. A replaceable sensing cartridge can further reduce downtime by avoiding sensor disassembly, electrical disconnection, optical realignment, and recalibration.

This paper describes a compact, rounded vision-based tactile finger designed for improved durability and rapid field replacement. Its sensing surface consists of a soft silicone gel with a nonpigmented, textured thin thermoplastic-polyurethane protective film and a removable polymer cartridge. The construction uses commercially available materials and a fabrication process intended to be suitable for economical manufacture.

Two accelerated comparative tests were developed to evaluate the sensing surface. A rotating-drum sanding test measured resistance to sustained abrasive wear, while a repetitive probe test evaluated durability under concentrated normal loading accompanied by lateral displacement. The developed sensor was compared under the same laboratory conditions with commercially available GelSight Mini \cite{gelsight2026mini} and DIGIT \cite{lambeta2020digit} sensors. In addition to test lifetime, the study examined the progression of physical damage, preservation of tactile imaging after damage, and the practicality of replacing the sensing cartridge.

The principal contributions of this work are: (1) a substantially more durable protective sensing surface under the accelerated test conditions used here; (2) gradual failure behavior that preserves useful tactile imaging after localized damage begins; (3) a rapidly replaceable sensing cartridge requiring no electrical or optical reconnection; and (4) defined accelerated procedures for comparing abrasion and repetitive-contact durability among vision-based tactile sensor constructions.


\FloatBarrier
\section{Hardware}
\label{sec:hardware}

\begin{figure}[htbp] 
    \centering
    \includegraphics[width=0.75\linewidth]{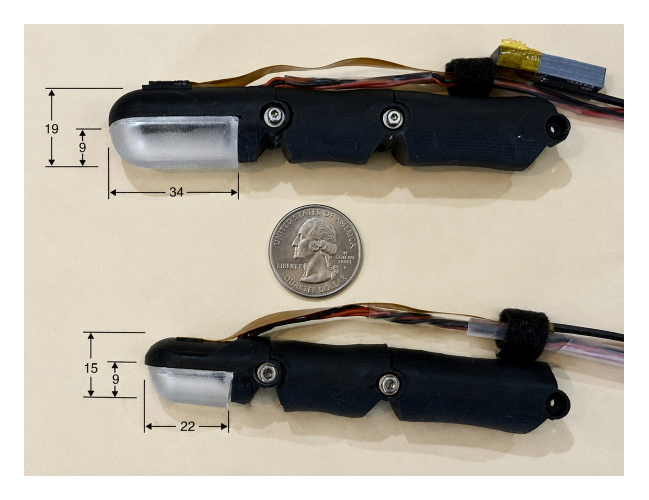}
    \vspace{-10pt}
    \caption{Humanoid vision-based tactile robot fingers.}
    \label{fig:humanoid_vbts}
\end{figure}

Two rounded vision-based tactile fingers were developed as interchangeable sensor platforms (\autoref{fig:humanoid_vbts}). They differ principally in fingertip size; the smaller design more closely approximates the dimensions of a human finger and was used for most experiments reported here.

The finger consists of four principal components: a miniature color camera, RGB LED illumination boards, a rigid finger housing, and a removable sensing cartridge. The cartridge snaps onto the housing and contains the compliant vision-based tactile sensor. This modular construction allows damaged sensors to be replaced rapidly without disassembling the vision-based hardware.

\begin{figure}[htbp] 
    \centering
    \includegraphics[width=0.75\linewidth]{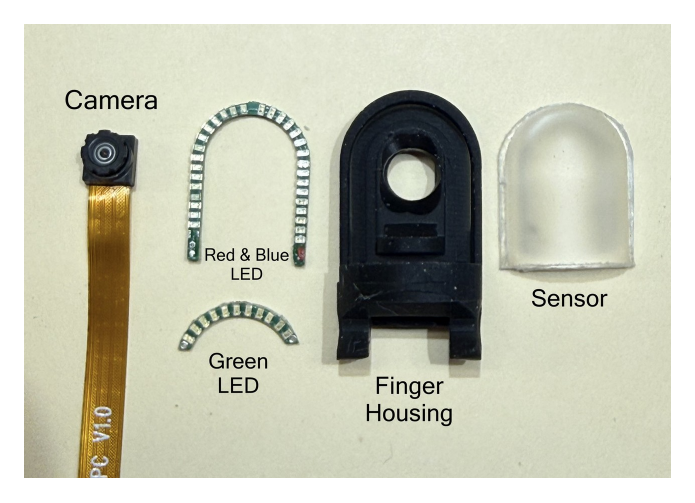}
    \vspace{-10pt}
    \caption{The components of the humanoid vision-based tactile finger.}
    \label{fig:components_humanoid_vbts}
\end{figure}

\begin{figure}[htbp] 
    \centering
    \includegraphics[width=0.75\linewidth]{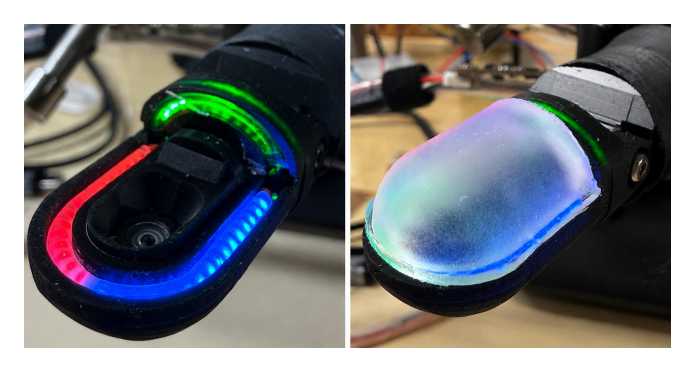}
    \vspace{-10pt}
    \caption{Assembly of the finger housing with the LEDs and the camera, (right side) the sensor is snapped onto the finger housing.}
    \label{fig:assembly_of_housing}
\end{figure}

\autoref{fig:assembly_of_housing} (left side) shows the assembly of the finger housing with the camera and the two LED boards with the LEDs illuminated. \autoref{fig:assembly_of_housing} (right side) shows the finger with the cartridge inserted (snap on) over the finger housing. 

The humanoid finger platform was intentionally designed to accept interchangeable sensing cartridges, allowing evaluation of multiple vision-based tactile sensing technologies without modification of the vision-based hardware.

\begin{figure}[htbp] 
    \centering
    \includegraphics[width=0.6\linewidth]{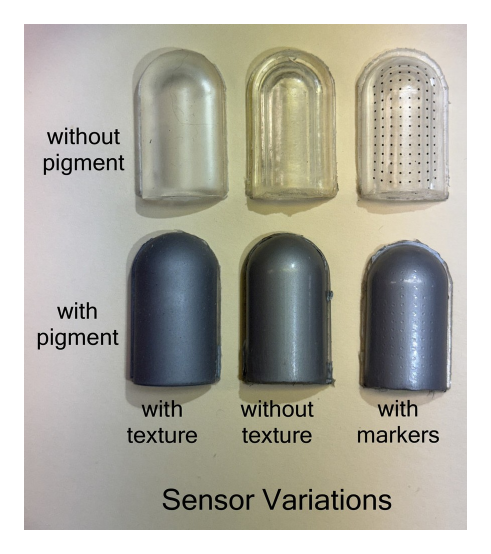}
    \vspace{-10pt}
    \caption{Examples of interchangeable vision-based tactile sensing cartridges compatible with the humanoid finger platform. The present work focuses on the transparent textured sensor (upper left).}
    \label{fig:sensor_variations}
\end{figure}

Several sensor architectures have been developed for use, including pigmented \cite{zhang2022hardware, abad2020visuotactile, yuan2017gelsight, lambeta2020digit, romero2020soft} and nonpigmented sensing surfaces \cite{yamaguchi2019tactile, hogan2021seeing, hogan2022fingersts, zhang2023tirgel}, smooth and textured outer surfaces, and sensors incorporating fiducial markers for force estimation. An example of each of these sensors is shown in \autoref{fig:sensor_variations}.

Among the various sensor architectures compatible with the humanoid finger platform, the transparent textured sensor was selected for this investigation because it provides excellent vision-based replication of surface detail while allowing the durability of the protective polyurethane film to be evaluated independently of pigment layers or fiducial markers. The durability improvements reported here are expected to be applicable to the other sensor architectures shown in \autoref{fig:sensor_variations} with only minor modifications.

\subsection{Sensor Imaging}
\label{subsec:Imaging}

\begin{figure}[htbp] 
    \centering
    \includegraphics[width=0.5\linewidth]{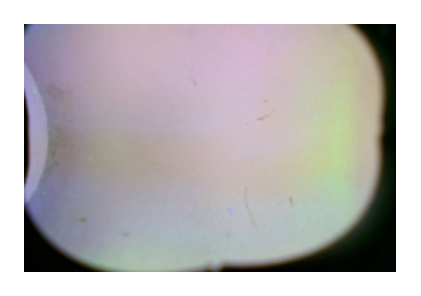}
    \vspace{-10pt}
    \caption{No contact image.}
    \label{fig:no_contact_image}
\end{figure}

\autoref{fig:no_contact_image} is an image from the sensor camera without sensor surface contact. The textured sensor surface blurs images coming to the sensor surface by light from outside. The internal RGB illumination is light piped through the sensor gel with some light scattered from the textured surface.

\begin{figure}[htbp] 
    \centering
    \includegraphics[width=1.1\linewidth]{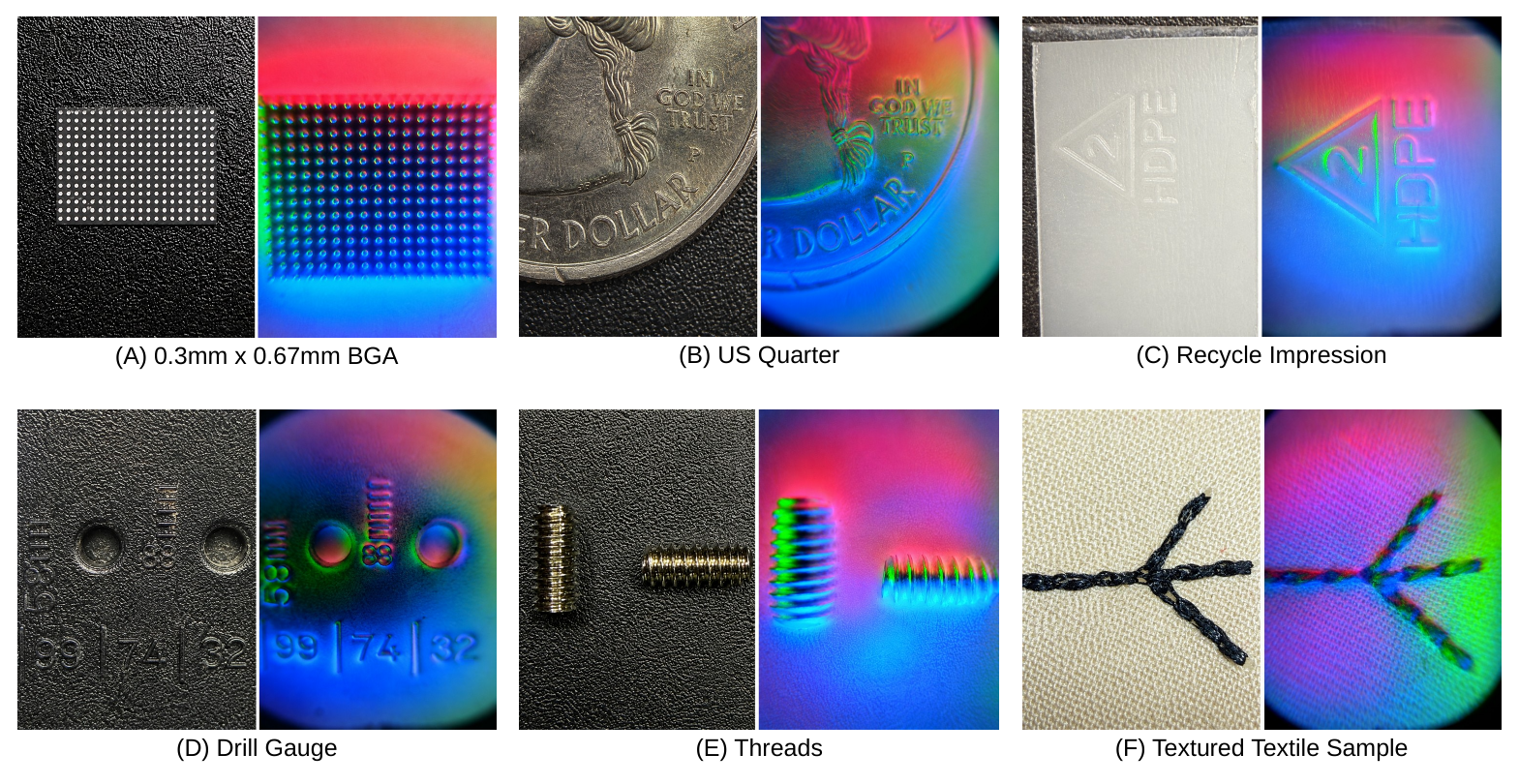}
    \vspace{-15pt}
    \caption{Left image - object photograph, right image - finger camera image}
    \label{fig:contact_objects}
\end{figure}

\autoref{fig:contact_objects} demonstrates the imaging capability of the transparent textured sensor. Fine surface features including BGA balls, embossed lettering on a U.S. quarter, molded recycling symbols, threaded fasteners, drilled gauge holes, and textile stitching are reproduced with high spatial fidelity. The soft silicone gel (Shore A 6.5) was selected to preserve the compliance required for sensitive tactile imaging.

\FloatBarrier
\section{Materials}
\label{sec:Materials}

\subsection{Design Requirements Driving Chemistry Selection}
\label{subsec:designrequirements}
The vision-based tactile sensor (VBTS) consists of a compliant optically clear elastomeric gel, a mechanically durable outer skin, and a rigid cartridge substrate. During operation, the sensor may be subjected to repetitive normal and shear loading, surface abrasion, large elastic deformation, and interfacial stresses that can cause surface rupture or delamination \cite{zhang2023tirgel, davis2025benchmarking}.

The combined gel and outer skin are required to satisfy five simultaneous requirements:
\begin{itemize}
    \item High optical transparency.
    \item Resistance to repetitive abrasion.
    \item Rapid elastic recovery with minimal compression set.
    \item Durable covalent bonding between the protective skin, silicone gel, and rigid cartridge.
    \item Compatibility with a simple, reproducible manufacturing process.
\end{itemize}

Final material decisions were guided by controlled abrasion testing and repetitive probe testing as described elsewhere in this work. 

\subsection{Gel Material Selection}
\label{subsection:gel_material_section}

Two gel families were evaluated: TPE-S (styrenic block copolymers) and platinum-cured silicone. Although TPE-S provided good optical clarity, repetitive touch testing demonstrated unacceptable compression set and permanent surface deformation. Platinum-cured silicone exhibited excellent elastic recovery and negligible compression set under identical testing conditions and was therefore selected despite its relatively low tear strength. The low tear strength was addressed through the use of a durable polyurethane protective film. Previous VBTS designs have similarly employed replaceable latex or polymer films to protect compliant sensing surfaces from wear \cite{rayamane2022design, zhao2025polytouch}.

\subsection{Silicone Gel}
\label{subsection:silicone_gel}
Silicone, Inc., XP565 was selected as the base material. This formulation provided the desired combination of optical clarity, compliance, and elastic recovery for tactile sensing. The gel was formulated at a non-standard 15:1 (A:B) ratio to reduce hardness. The resulting gel exhibited a measured Shore A hardness of 6.5. 

\subsection{Skin Material Selection}
\label{subsection:skin_material_selection}
Several protective skin materials were evaluated, including TPE-S films, polyvinyl alcohol, and thermoplastic polyurethanes. TPU provided the best combination of abrasion resistance, elasticity, optical transparency, and chemical durability and was selected for subsequent development.

\subsection{Thermoplastic Polyurethanes}

Thermoplastic polyurethanes (TPU) provided a compelling combination of superior abrasion resistance, optical transparency, high elasticity, chemical resistance, and a broad hardness range within compatible chemical families. Although TPU requires additional coupling chemistry (e.g., silane-based systems) to achieve durable bonding to silicone gels, its mechanical durability and absence of compression set was significantly superior to TPE-S systems.

Early development focused on mechanically graded multilayer TPU structures intended to reduce interfacial shear stress. Although these structures performed well, subsequent testing demonstrated that six sequential coatings of a single TPU grade (BASF Elastollan 1170A10, Shore A 70) provided comparable durability while substantially simplifying manufacturing. The simpler architecture was therefore adopted for all subsequent work.

An important outcome of this development was that the final sensor architecture proved substantially simpler than initially anticipated. Rather than requiring a mechanically graded multilayer structure, high durability was achieved using a single polyurethane chemistry applied as six sequential coatings. The resulting architecture consists of a single TPU chemistry, a single silicone formulation, and a reproducible bonding process using commercially available materials. This simplicity is expected to facilitate both laboratory fabrication and manufacturing.

During development, the abrasion resistance of Elastollan 1170A10 coatings was found to depend strongly on thermal conditioning of the coating solution before deposition. The optimized conditioning procedure increased sanding lifetime by approximately a factor of two to three and was therefore incorporated into the final fabrication process. 

\subsection{Interfacial Bonding}
Durable service life required strong interfacial bonding among the cartridge, silicone gel, and TPU protective skin. Corona activation followed by silane coupling treatment provided the required adhesion and was therefore incorporated into the final manufacturing process.

\subsection{Cartridge Material}
The cartridge was injection molded from Kuraray Septon 2104, which provides adequate rigidity for structural support while retaining sufficient compliance for snap-in replacement.

\FloatBarrier
\section{Sensor Preparation}

\subsection{Sensor Overview}
\begin{figure}[htbp] 
    \centering
    \includegraphics[width=1\linewidth]{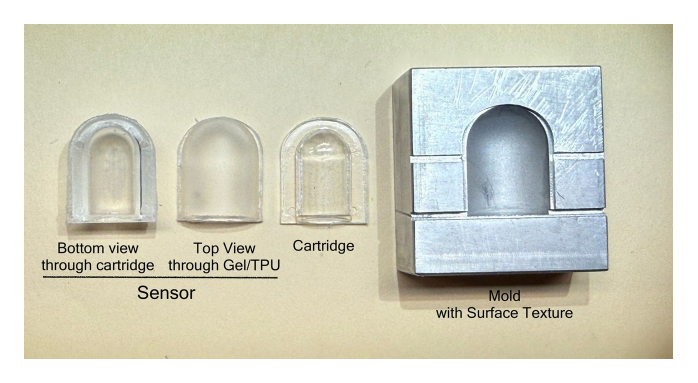}
    \vspace{-20pt}
    \caption{Sensor manufacturing components (mold and cartridge) and completed sensor. }
    \label{fig:manufacturing_components}
\end{figure}

The vision-based tactile sensors used in this study were fabricated using a textured aluminum mold and a removable polymer cartridge (\autoref{fig:manufacturing_components}). The completed sensor consisted of an approximately 85\,\textmu{}m polyurethane protective film bonded to a 3 mm thick platinum-cured silicone gel (Shore A $\approx$ 6), which was in turn bonded to a rigid polymer cartridge. The textured mold surface defined the final sensing surface of the sensor. 

The fabrication process consisted of mold preparation, polyurethane film deposition, cartridge surface preparation, silicone gel casting, and final demolding.

\subsection{Mold preparation}
The sensor mold was machined from aluminum and contained the surface texture that was replicated onto the finished tactile sensing surface. Before coating, the mold was thoroughly cleaned and treated with a thin layer of undiluted McLube 1805 mold release to ensure consistent sensor removal while preserving the fidelity of the replicated surface texture.

\subsection{Polyurethane protective film}
The protective outer film was fabricated from BASF Elastollan 1170A10 polyurethane using a 7 wt.\% TPU solution in 80 wt.\% tetrahydrofuran (THF) and 20 wt.\% methyl ethyl ketone (MEK) as solvent. The coating solution was thermally conditioned by cooling it to 60\textdegree{}F (16\textdegree{}C) until a fine haze developed and then warming it to approximately 90\textdegree{}F (32\textdegree{}C) before coating. Six sequential wet-on-wet coatings were applied, allowing partial solvent evaporation between applications. Five molds were coated sequentially, resulting in approximately 4 minutes between successive coating applications on each mold. Following deposition of the final layer, the coated mold was oven dried at 150\textdegree{}F for 30 minutes and subsequently vacuum dried at the same temperature for an additional 30 minutes to minimize residual solvent. Final film thickness was approximately 85\,\textmu{}m.

\subsection{Cartridge Surface Preparation}
The polymer cartridge was corona treated to increase surface energy and subsequently brush-coated with a thin layer of undiluted Momentive SS4120 silane coupling agent to promote adhesion between the cartridge and the silicone gel. The silane layer was allowed to react for 30 minutes at 50\% relative humidity and 80\textdegree{}F before thermal curing at 150\textdegree{}F for 30 minutes.

\subsection{Coated Mold Surface Preparation}
The exposed TPU surfaces on coated molds were lightly corona-treated before application of the silane coupling agent. These treatments were intended to promote adhesion between the polyurethane film and the silicone gel.  As with the cartridges the silane was allowed to react under controlled humidity conditions before thermal curing. 

\subsection{Silicone Gel Casting}
The mold cavity was filled with a platinum-cured silicone gel having a Shore A hardness of approximately 6. Silicone Inc. XP565 was used at an A:B mixing ratio of 15:1. The prepared cartridge was inserted into the uncured gel, displacing excess material while accurately positioning the cartridge within the mold. The completed assembly was cured at 180\textdegree{}F (82\textdegree{}C) for 3.5 hours.

\subsection{Demolding}
Following curing, the completed sensor was removed from the mold. The textured mold surface was replicated directly into the polyurethane film, producing the final tactile sensing surface without additional processing.


\FloatBarrier
\section{Testing}
\label{sec:testing}

The objective of this work was to develop a vision-based tactile sensor capable of tolerating repeated contact and surface abrasion in demanding robotic applications. Two complementary accelerated durability tests were used. A rotating-drum sanding test evaluated resistance to sustained abrasive wear, while a repetitive probe test (RPT) evaluated durability under repeated concentrated normal loading accompanied by lateral displacement.

No broadly accepted durability standard has been established specifically for layered vision-based tactile sensors. The procedures developed here therefore used defined geometries, loads, cycling conditions, and failure criteria to permit comparisons among sensor constructions under common laboratory conditions. The tests were designed to accelerate selected damage mechanisms rather than reproduce a particular industrial task or predict field service life.

\subsection{Repetitive Probe Test (RPT) Apparatus}
The RPT subjected the sensor to repeated concentrated normal loading accompanied by lateral probe motion. This mixed-mode loading was selected to accelerate surface deformation, interfacial delamination, probe-imprint formation, and other damage associated with repeated high-strain contact. It was intended as a comparative endurance test rather than a direct simulation of a particular robotic operation.

\begin{figure}[htbp] 
    \centering
    \includegraphics[width=0.8\linewidth]{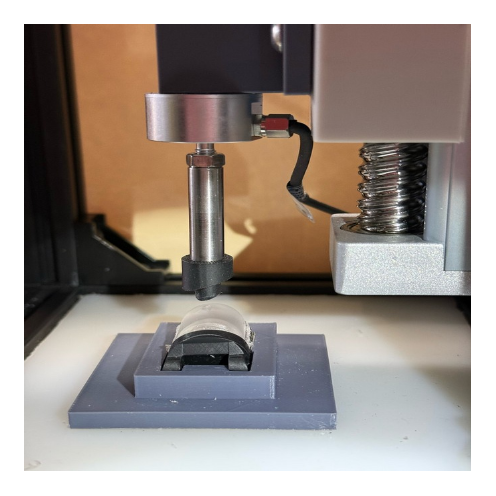}
    \vspace{-15pt}
    \caption{Repetitive probe test (RPT) apparatus.}
    \label{fig:rpt_apparatus}
\end{figure}

The RPT apparatus consisted of a rigidly mounted finger sensor positioned beneath a vertically actuated probe. The probe was attached to a load cell mounted on a linear carriage driven by a lead screw and stepper motor (\autoref{fig:rpt_apparatus}). Interchangeable 3D-printed probe tips were used. Unless otherwise stated, testing employed a cylindrical probe 6 mm in diameter with radiused edges.

As shown in \autoref{fig:rpt_apparatus}, the finger sensor is mounted in a rigid holder beneath a 6-mm-diameter probe attached to a 10 kgf load cell. The probe was actuated by a stepper–motor driven lead screw. Lateral compliance within the carriage assembly produces measurable lateral probe displacement under high load conditions.

The probe axis was mounted vertical with the contacting face angled 20\textdegree{}. Before each test, the probe position was adjusted until the load cell indicated the target peak force. The actuator stroke was then held constant during cycling, and peak force was checked periodically. The probe was held at peak force for 0.3 s and remained unloaded for 0.5 s, including the unloading motion. Testing was conducted at 45 cycles per minute and commonly continued 24 h/day for extended durations. 

\subsection{Probe Penetration and Lateral Displacement}

\begin{figure}[htbp] 
    \centering
    \includegraphics[width=1\linewidth]{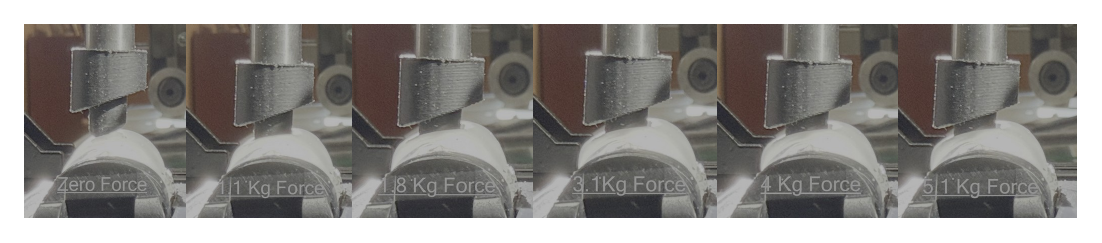}
    \vspace{-15pt}
    \caption{Probe position as a function of applied load for the 6-mm-diameter probe having a contact face angled at 20\textdegree{}.}
    \label{fig:probe_position}
\end{figure}

To quantify probe penetration depth and lateral displacement under load, a series of images were captured at incrementally increasing applied forces. From these images (\autoref{fig:probe_position}), both vertical penetration and lateral probe tip displacement were determined. Images included a calibrated length reference, and probe positions were measured relative to the unloaded position.

\begin{figure}[htbp] 
    \centering
    \includegraphics[width=1\linewidth]{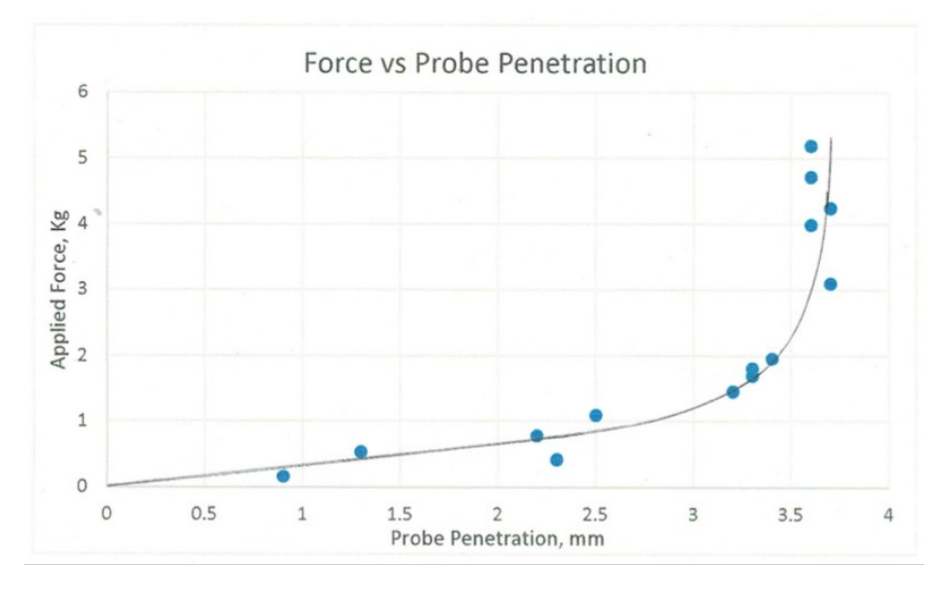}
    \vspace{-15pt}
    \caption{Probe force as a function of penetration depth into the tactile sensor. A rapid increase in force is observed as penetration approaches approximately 3.5 mm, corresponding to near – complete compression of the 3.5 mm gel thickness. The non-linear increase reflects geometric confinement of the gel against the underlying cartridge.}
    \label{fig:probe_force_depth}
\end{figure}

Force–penetration measurements showed a rapid increase in load as penetration approached approximately 3.5 mm (\autoref{fig:probe_force_depth}). The nominal gel thickness was 3.0 mm. Owing to the curved sensor geometry, the measured local gel thickness at the slightly off-center probe-contact position was approximately 3.5 mm. At high loads, the probe therefore approached the underlying cartridge.

\begin{figure}[htbp] 
    \centering
    \includegraphics[width=1\linewidth]{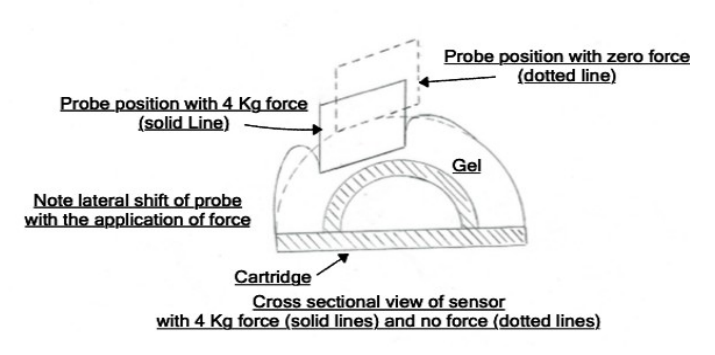}
    \vspace{-15pt}
    \caption{Sectional representation of probe position at 39.2 N (4.0 kgf) applied force. The remaining separation between the probe surface and the underlying cartridge is less than approximately 0.5 mm. Under this condition lateral probe displacement of approximately 1.1 mm is observed producing a confined shear deformation state.}
    \label{fig:side_view}
\end{figure}

Under a 39.2 N (4 kgf) applied force, the following conditions were observed (\autoref{fig:side_view}):

\begin{enumerate}
    \item Probe penetration approached approximately 3.5\,mm.
    \item The minimum separation between the probe and underlying cartridge was less than approximately 0.5\,mm.
    \item Lateral probe displacement was approximately 1.1\,mm.
\end{enumerate}

\begin{figure}[htbp] 
    \centering
    \includegraphics[width=1\linewidth]{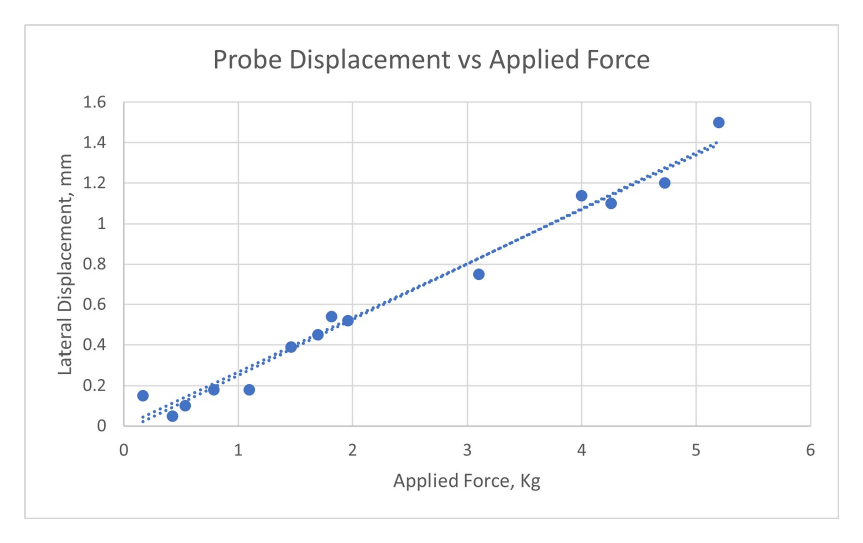}
    \vspace{-15pt}
    \caption{Lateral displacement of the probe tip as a function of applied normal force. The approximately linear relationship reflects compliance within the load-cell carriage assembly. This lateral motion imposes additional cyclic shear deformation on the sensor during RPT operation.}
    \label{fig:lateral_displacement_plot}
\end{figure}

The lateral displacement arises from compliance within the load-cell cartridge assembly. This displacement (\autoref{fig:lateral_displacement_plot}) was repeatedly observed and remained approximately stable during extended testing.

\subsection{Shear Confinement Under High Loads}
At high penetration depths, the gel layer becomes confined between the probe and the underlying cartridge. Under these conditions, the ratio of lateral probe displacement to the minimum remaining compliant thickness exceeded approximately 2. This geometric ratio indicates severe confined shear deformation, although it should not be interpreted as a direct measurement of local material strain.

These conditions represent a deliberately severe loading configuration. By forcing lateral deformation into a reduced compliant thickness, the RPT imposes high interfacial shear and concentrated deformation within the skin–gel and gel-cartridge transition zones.

The RPT protocol was therefore intended as a deliberately severe accelerated endurance test rather than a direct simulation of typical service contact. It provides a comparative means of evaluating interfacial durability under concentrated mixed-mode loading.

\subsection{Abrasion Resistance Characterization (Sanding Test)}

The rotating-drum sanding test was developed to compare resistance to sustained sliding abrasion and to characterize the morphology of protective-film failure. The procedure was intended as an accelerated comparative test rather than a simulation of a specific industrial contact.

\begin{figure}[htbp] 
    \centering
    \includegraphics[width=0.90\linewidth]{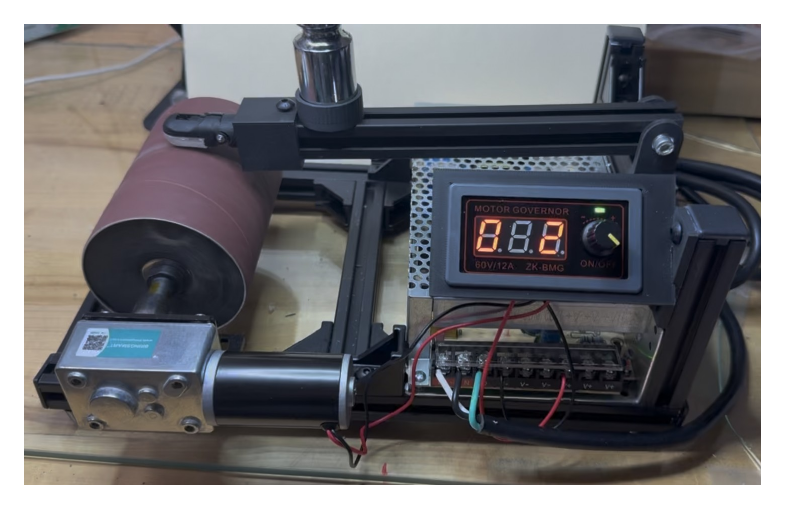}
    \vspace{-10pt}
    \caption{Rotating-drum abrasion apparatus used to evaluate surface durability of the TPU stack. The 3 inch diameter drum is wrapped with 400 grit abrasive paper and operated at 80 rpm. Normal force is applied via calibrated weights on pivot arm assembly.}
    \label{fig:rotating_drum_abrasion}
\end{figure}

Abrasion resistance was evaluated using a custom rotating-drum apparatus (\autoref{fig:rotating_drum_abrasion}). The apparatus consisted of a 3-in. (76-mm)-diameter cylindrical drum wrapped with 400-grit abrasive paper and operated at 80 rpm. The sensor was mounted on a pivoting arm and pressed against the rotating abrasive surface. Applied load was adjusted using weights on the pivot arm, and the effective normal force at the sensor contact position was verified with a digital scale before testing.

Tests used digital-scale weight readings of 100, 200, 300, or 400 g, corresponding to normal forces of approximately 0.98, 1.96, 2.94, and 3.92 N, respectively. Unless otherwise stated, abrasion performance was evaluated at a scale reading of 200 g, corresponding to approximately 1.96 N.

Sanding progression was monitored by direct observation of the sensor surface during testing, and surface conditions were normally recorded at intervals of 5 or 10 min. The inspection frequency was increased as visible wear approached the rupture endpoint. Testing continued until the first visually detectable rupture of the TPU protective film exposed the underlying silicone gel. This event was defined as the abrasion endpoint, and elapsed sanding time was recorded in minutes. A fresh sandpaper surface was used for each test. The pivot arm holding the cartridge can be adjusted to move the sensor to a fresh position along the sanding drum for each test. The sandpaper was used as received.

The sanding test is intended to provide a comparative abrasion metric under controlled and repeatable conditions rather than a replicate of a specific real-world surface. The rotating drum configuration produces sustained sliding contact and accelerated wear, thereby providing a conservative assessment of surface durability.

\subsection{DIGIT and Mini Testing}
Commercially available GelSight Mini \cite{gelsight2026mini} and DIGIT \cite{lambeta2020digit} cartridges were evaluated using the same sanding apparatus, abrasive paper, drum speed, and measured normal load used for the developed sensors. A separate 3D-printed fixture was prepared for each commercial cartridge to position its sensing surface against the rotating drum. The fixtures also permitted the sensor surface to be observed through the gel during testing. Each test was recorded by video, allowing the time of initial rupture of the pigmented surface layer to be determined from the recording.

Separate 3D-printed fixtures were also prepared for RPT of the Mini and DIGIT cartridges. Because both commercial sensing surfaces were flat, each cartridge was oriented at 20\textdegree{} to the horizontal so that its surface was aligned with the angled contact face of the probe. The same probe, peak load, cycle rate, and actuator procedure used for the developed sensors were employed. Testing was interrupted at 5-min intervals to photograph the sensing surfaces and document damage progression.

Three GelSight Mini Standard Silicone Replacement Gels (serial nos. 8563–8565) and three DIGIT Replacement Cartridges (serial nos. 2376–2378) were purchased directly from GelSight Inc. on March 27, 2026, for comparative testing. The specimens were tested as received without surface modification or pretreatment. One Mini replacement gel and one DIGIT replacement cartridge were evaluated using the sanding test, and separate specimens of each product were evaluated using the RPT. 

\FloatBarrier
\section{Results}
\label{sec:results}

\subsection{Abrasion resistance}
\begin{figure}[htbp] 
    \centering
    \includegraphics[width=1\linewidth]{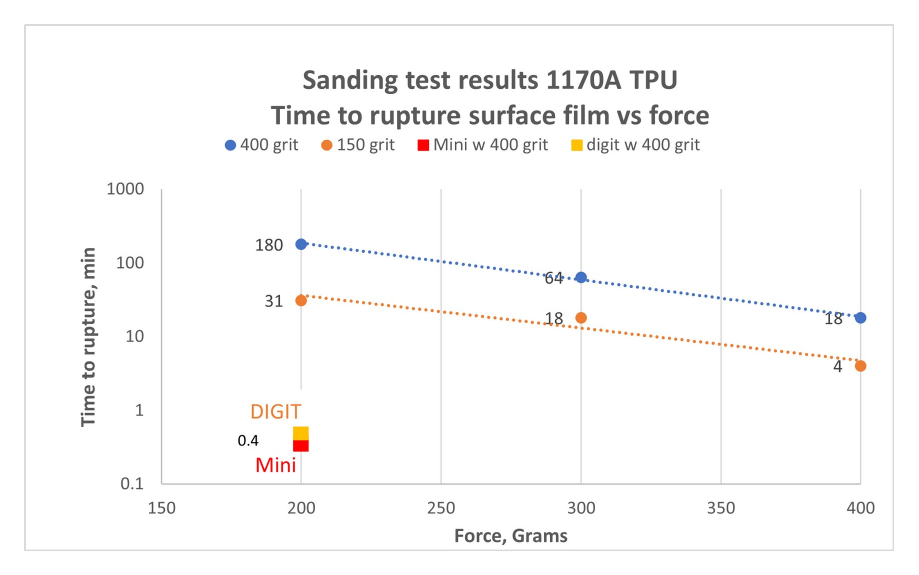}
    \vspace{-10pt}
    \caption{Sanding results.}
    \label{fig:sanding_results_plot}
\end{figure}

Abrasion resistance was evaluated using the rotating sandpaper test described previously. Table 1 summarizes the time required for rupture of the TPU protective film for different sandpaper grits and applied loads. These data are plotted in \autoref{fig:sanding_results_plot}.

\begin{table}[htbp]
    \centering
    \begin{threeparttable}
    \caption{Sanding-test time to first protective-film rupture, in minutes.}
    \label{tab:sanding}
    \begin{tabular}{ccccc}
        \toprule
        Digital-scale & \multicolumn{2}{c}{Developed sensor} & DIGIT & Mini \\
        \cmidrule(lr){2-3} \cmidrule(lr){4-4} \cmidrule(lr){5-5}
        reading (g) & 400 grit & 150 grit & 400 grit & 400 grit \\
        \midrule
        200 & 180 ($n = 2$) & 31 ($n = 2$) & 0.5 ($n = 1$) & 0.4 ($n = 1$) \\
        300 & 64 ($n = 1$)  & 18 ($n = 2$) & ---   & ---  \\
        400 & 18 ($n = 2$)  & 4 ($n = 2$)  & ---           & ---           \\
        \bottomrule
    \end{tabular}
    \begin{tablenotes}[flushleft]
        \footnotesize
        \item Note: Values are mean times to first visually detectable protective-film rupture, in minutes; n = number of specimens.
    \end{tablenotes}
    \end{threeparttable}
\end{table}

At the standard test condition of 400-grit paper and a 200 g scale reading, the mean time to protective-film rupture was 180 minutes (n = 2; range, 176–184 minutes). Increasing the applied force reduced the mean lifetime to 64 minutes (n = 1) at 300 g and 18 minutes (n = 2) at 400 g. With 150-grit paper, the corresponding mean lifetimes were 31, 18, and 4 minutes at loads of 200, 300, and 400 g, respectively (n = 2 for each condition).

Under the same standard test conditions, the commercial DIGIT and Mini sensors failed after approximately 0.5 and 0.4 minutes respectively. 

\autoref{fig:sanding_results_plot} illustrates the strong dependence of abrasion lifetime on applied load. Lifetime decreased approximately tenfold as the normal force increased from 200 g to 400 g. Despite this severe loading, the developed sensor substantially outperformed the commercial reference sensors.

\begin{figure}[htbp] 
    \centering
    \includegraphics[width=1.1\linewidth]{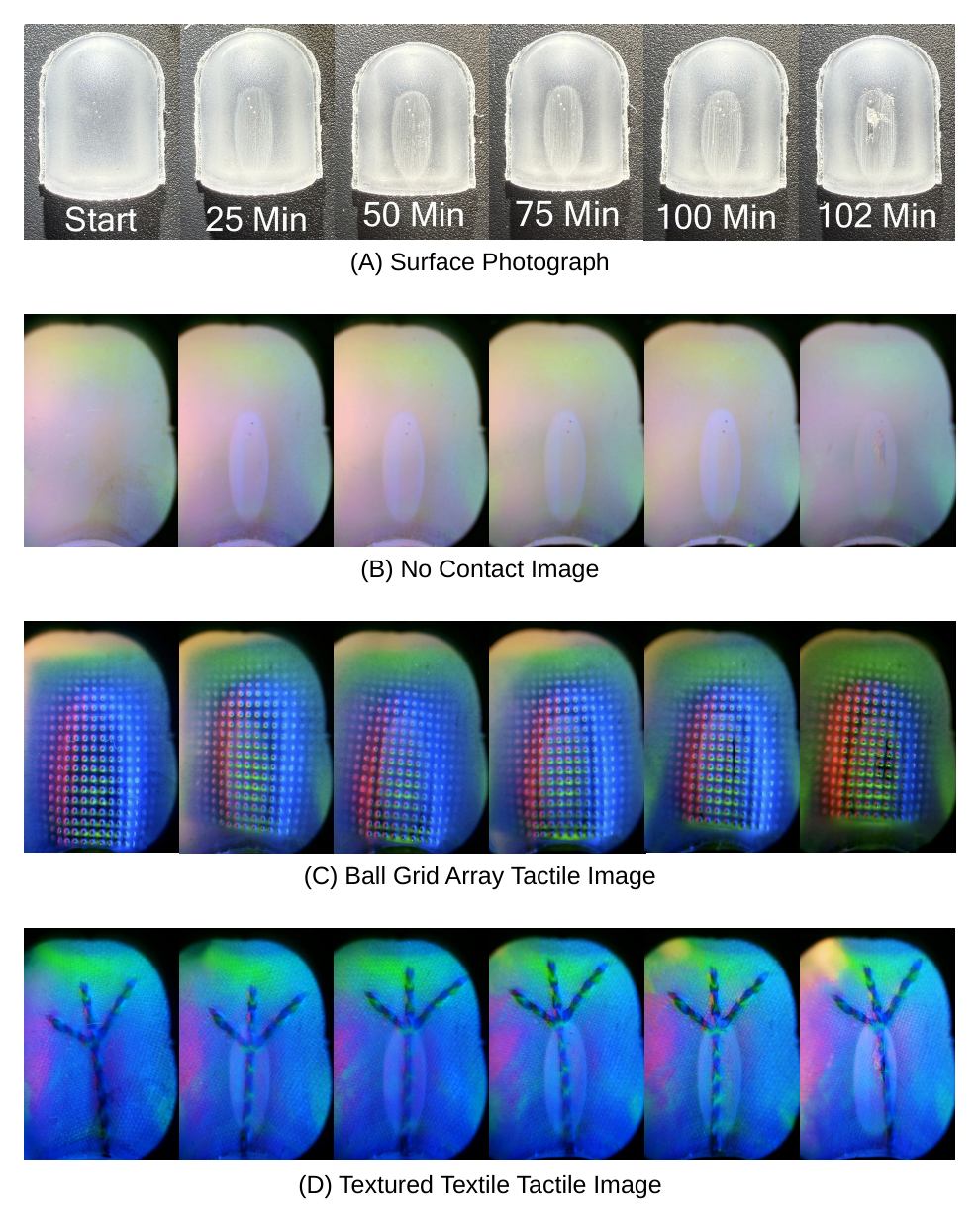}
    \vspace{-20pt}
    \caption{The results of a sanding test. Surface photographs and tactile sensor photographs vs time during sanding test. (The specimen shown is representative of the progression of wear and was not one of the two specimens used to calculate the 180-minute value in Table 1.)}
    \label{fig:sanding_results_pics}
\end{figure}

As shown in the surface photographs in \autoref{fig:sanding_results_pics}, abrasion produced progressively increased light scattering within the worn region. Despite the localized change in optical appearance, tactile images of the contacted objects retained useful detail throughout the test.

\begin{figure}[htbp] 
    \centering
    \includegraphics[width=1\linewidth]{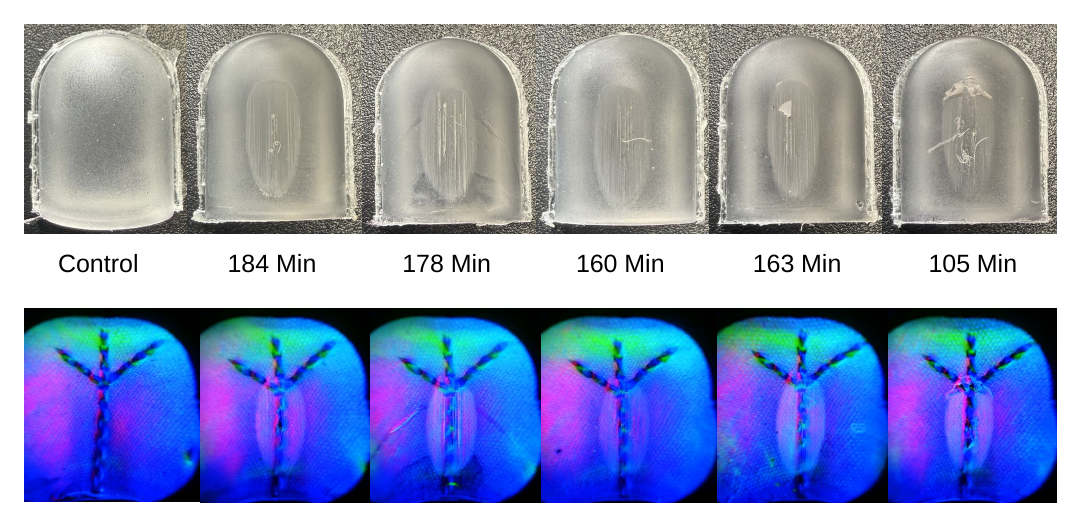}
    \vspace{-20pt}
    \caption{Terminal sanding images - Top: Surface photos with time of sanding, Below: tactile camera images of textured textile sample.}
    \label{fig:terminal_sanding_images}
\end{figure}

\autoref{fig:terminal_sanding_images} shows five representative tested sensors and an untested control. Although rupture time varied among the specimens, the wear morphology was consistent. Failure consisted of localized rupture of the TPU protective film rather than catastrophic loss of the sensing surface. At the defined endpoint, most of the sensing area remained intact and tactile images retained useful detail. Thus, the endpoint represents the onset of protective-film rupture rather than complete functional failure.

\begin{figure}[htbp] 
    \centering
    \includegraphics[width=1\linewidth]{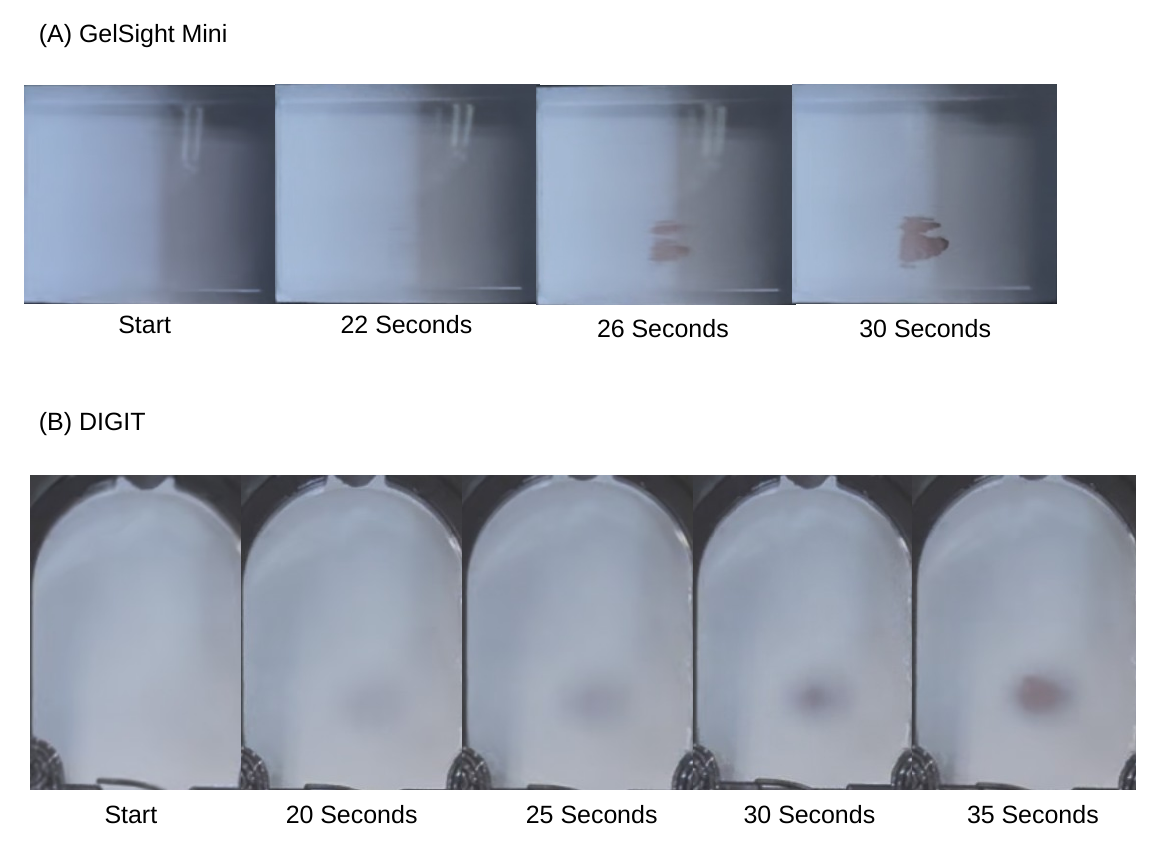}
    \vspace{-10pt}
    \caption{GelSight Mini and DIGIT Sanding Test - photographed through the sensor during testing (200 gram force, 400 grit sandpaper, 80 rpm on sanding drum, 3 inch diameter drum)}
    \label{fig:gelsight_digit_sanding}
\end{figure}

Under the same standard test conditions, commercially available DIGIT  and GelSight Mini  sensors reached the defined abrasion endpoint after approximately 30 and 24 seconds, respectively. Testing was continued briefly beyond this endpoint to document the progression of damage shown in \autoref{fig:gelsight_digit_sanding}.

\subsection{Repetitive Probe Testing}
\begin{figure}[htbp] 
    \centering
    \includegraphics[width=1.1\linewidth]{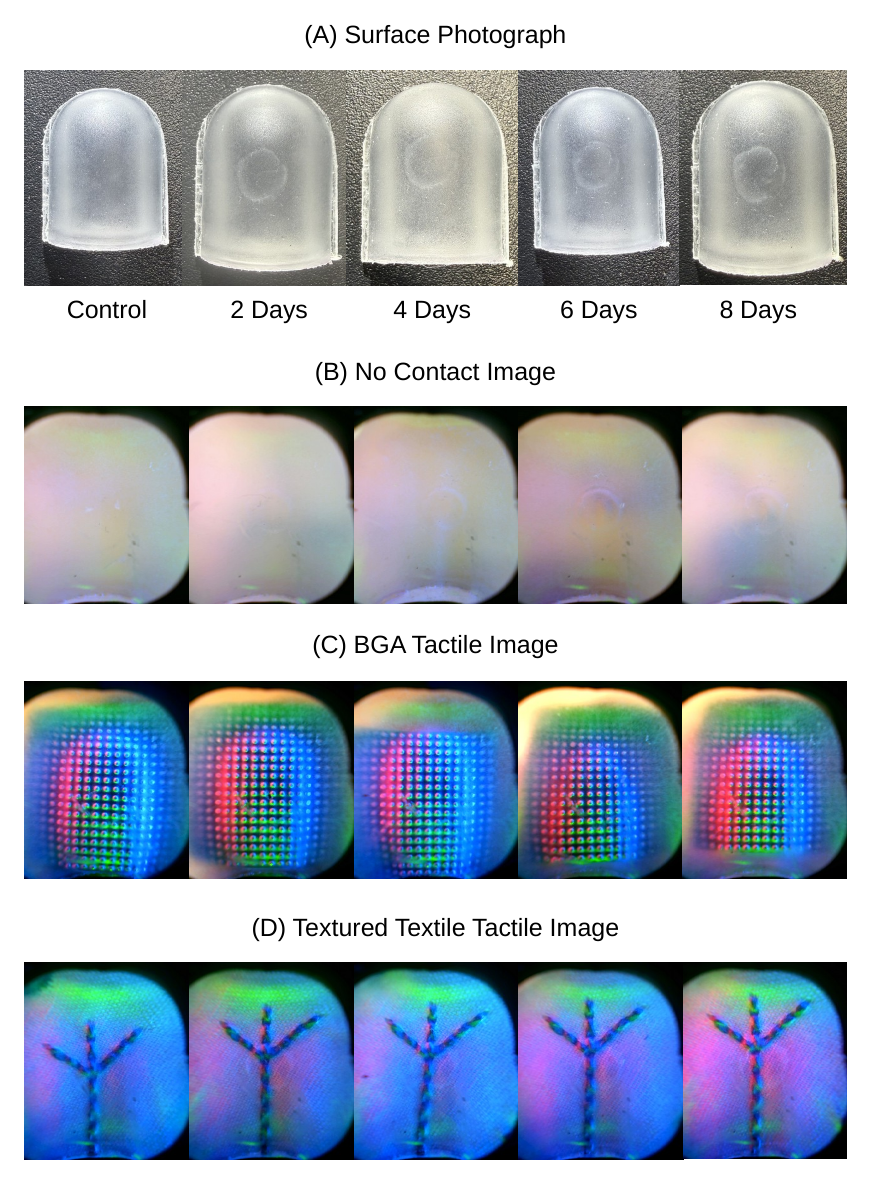}
    \vspace{-10pt}
    \caption{Repetitive Probe Test Results.}
    \label{fig:figure18}
\end{figure}

\autoref{fig:figure18} presents the condition of a representative sensor during eight days of continuous repetitive probe testing (520,000 contacts of a 6-mm-diameter probe with a force of 39 N (4 kgf)), together with tactile images acquired at selected intervals. A faint circular imprint corresponding to the probe circumference developed progressively on the sensor surface. Tactile images of the ball grid array and the textured textile sample changed little throughout the eight-day test. 

\begin{figure}[htbp] 
    \centering
    \includegraphics[width=1\linewidth]{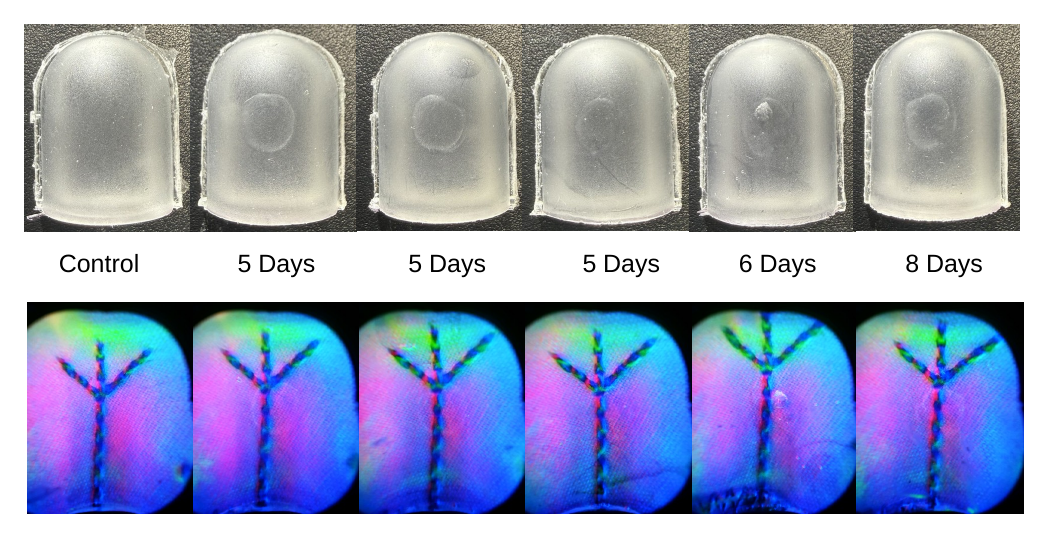}
    \vspace{-20pt}
    \caption{Terminal test images of repetitive probe test.}
    \label{fig:figure19}
\end{figure}

\autoref{fig:figure19} shows terminal surface photographs and tactile images from five sensors prepared in separate batches. Most sensors were tested continuously for five days. A circular imprint corresponding to the circumference of the reciprocating probe was visible on all tested sensors. The imprint produced little detectable interference with tactile imaging of the textured textile sample.

A less frequent failure mode was a localized blister formation caused by delamination of the TPU protective layer from the underlying silicone gel near the circumference of the probe contact region. The six-day specimen in \autoref{fig:figure19} exhibits an approximately 1-mm-diameter blister. The defect is visible both in the surface photograph and in the corresponding tactile image, although the remainder of the tactile image remains interpretable. These delamination defects were observed to grow slowly with continued testing and could ultimately interfere with tactile imaging. Under the present test conditions, localized TPU–gel delamination was the only failure mode observed to progress toward loss of useful tactile imaging. 

Nine sensors were evaluated using the current formulation. Seven were tested for five days, one for six days, and one for eight days. All nine developed a circular surface imprint corresponding to the circumference of the probe. This minor defect produced little detectable interference with tactile imaging. Localized TPU–gel delamination blisters developed in three sensors: two by the fifth day and one by the sixth day. These blisters remained small at the respective test endpoints and did not substantially impair tactile imaging. The sensor tested for eight days exhibited only the circular probe imprint and showed no evidence of delamination. All nine sensors therefore remained functionally usable through their respective test endpoints.

\subsection{Comparison with Commercial Vision-based Tactile Sensors}

\begin{figure}[htbp] 
    \centering
    \includegraphics[width=0.9\linewidth]{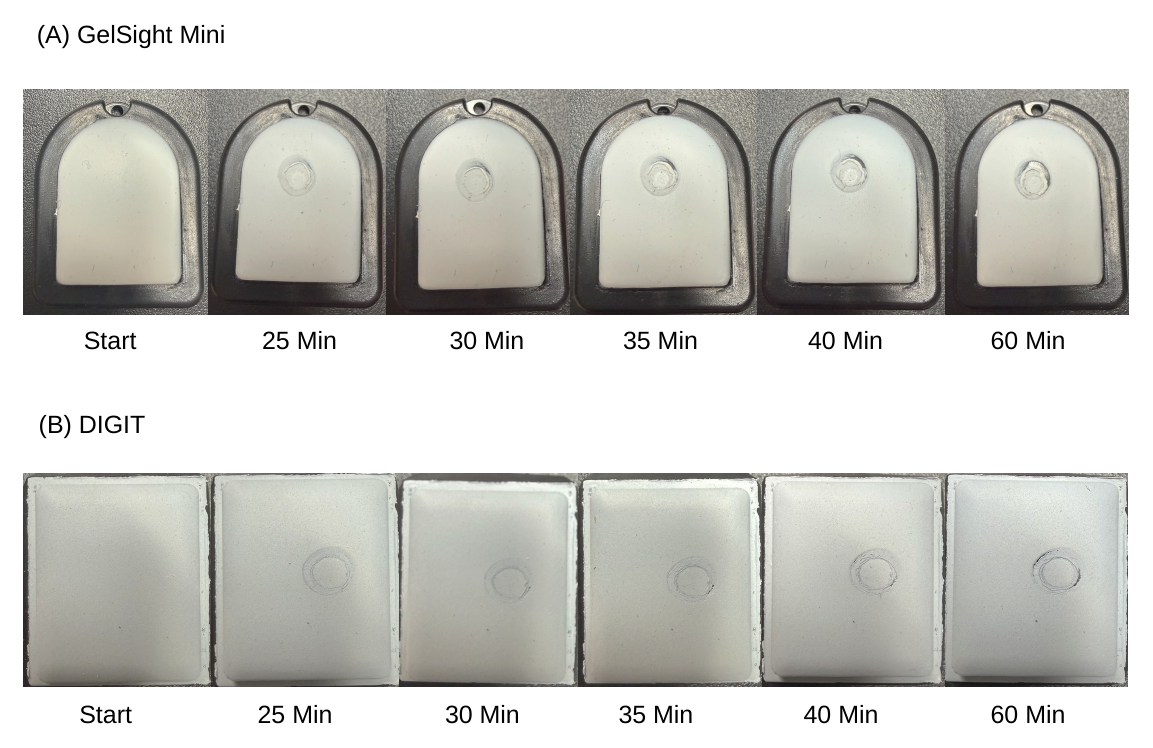}
    \vspace{-1pt}
    \caption{DIGIT and Mini RPT Results. Top surface photographs}
    \label{fig:digit_mini_rpt_test}
\end{figure}

The same repetitive probe test was applied to commercially available GelSight Mini and DIGIT sensors under the test conditions described above. As shown in \autoref{fig:digit_mini_rpt_test}, both commercial sensors developed initial rupture of the pigmented surface film within approximately 35 minutes and 25 minutes, respectively. The test was continued for 60 minutes to monitor damage progression. In comparison, all nine sensors produced using the present construction remained functionally usable through their respective test endpoints, which ranged from 5 to 8 days. Although all developed a minor circular probe imprint and three developed small localized delamination blisters, these defects produced little interference with tactile imaging. A comparison therefore indicates a large improvement in resistance to repetitive concentrated loading under this test protocol.

Because the commercial sensors differ from the present sensor in geometry, elastomer construction, protective-surface design, and intended operating conditions, these measurements should be interpreted as a comparison under a common laboratory test rather than as a complete assessment of overall sensor durability.

Both the DIGIT and Mini failed by surface abrasion along the imprinted image of the probe on the sensor surface. 

\FloatBarrier
\section{Discussion} 
The primary objective of this work was to develop a vision-based tactile sensor with improved durability and maintainability for demanding robotic applications. Under the defined accelerated laboratory tests, the developed sensor reached the protective-film rupture endpoint after approximately 2–3 hours of continuous abrasion. In repetitive probe testing, seven sensors remained functionally usable when testing was discontinued after 5 days, one remained usable when testing was discontinued after 6 days, and one remained usable when testing was discontinued after 8 days. Functional failure was therefore not observed within the respective test durations. Under the same laboratory protocols, commercially available GelSight Mini and DIGIT specimens exhibited initial surface-film rupture after approximately 24–30 seconds of sanding and 25–35 minutes of repetitive loading. These observations establish an improvement of more than two orders of magnitude under the defined accelerated conditions, although the full improvement in repetitive-loading lifetime could not be determined because the developed sensors had not reached functional failure. Because these tests were designed to accelerate selected damage mechanisms rather than reproduce a particular industrial operation, the results should not be interpreted as direct predictions of field service life.

These findings are consistent with recent work identifying polyurethane as a promising route to improved VBTS resilience and with PolyTouch, which demonstrated improved lifetime through a protected, replaceable tactile interface \cite{zhao2025polytouch, davis2025benchmarking}. Direct numerical comparison is not possible, however, because the loading geometries, forces, abrasion media, endpoints, and sensor constructions differ among the studies.

An equally important finding is the manner in which damage develops. For a tactile sensor used in a robotic system, functional lifetime depends not only on the onset of physical damage but also on whether useful tactile information remains available after damage begins. In the present sensors, abrasion first produces localized rupture of the TPU protective film while the majority of the sensing surface remains functional. Likewise, repetitive probe testing produces progressive probe imprints and, in some cases, localized delamination blisters that develop gradually rather than catastrophically. In the abrasion test, useful tactile images remained available beyond the defined protective-film rupture endpoint. During repetitive probe testing, circular imprints and small localized delamination blisters developed gradually, while the specimens remained functionally usable at their respective test endpoints.

Long sensor lifetime alone is insufficient for industrial deployment. Components that eventually wear must also be replaceable quickly and economically. The tactile sensing surface described here consists of a removable cartridge that can be replaced without sensor disassembly, electrical disconnection, optical realignment, recalibration, or specialized tools. Replacement requires only removal of the worn cartridge using a simple plastic prying tool, followed by installation of an identical replacement. The low component count and straightforward construction may also support economical manufacture and replacement. Together with the gradual failure behavior observed in this study, this architecture enables maintenance to be scheduled rather than performed as an emergency repair.

The sanding and repetitive probe tests also proved valuable as development tools. During sensor development, changes in film construction, interfacial preparation, environmental conditioning, and coating-solution preparation were associated with substantial changes in measured durability. Not all of these variables were investigated independently, and their individual contributions therefore cannot be quantified from the present study. Nevertheless, the tests provided sensitive comparative measures that helped identify promising fabrication conditions. The equipment required for both tests is inexpensive and readily available. Although the selected loading conditions were intended to accelerate damage rather than reproduce a particular industrial task, the methods provide defined procedures for comparing sensor constructions under common laboratory conditions. Additional interlaboratory testing and statistical characterization would be required before either procedure could be considered a standardized durability test.

Although this study focused on a nonpigmented vision-based tactile sensor, the durability concepts developed here are not limited to this particular optical architecture. The results of this study are applicable to a range of vision-based tactile sensors, including systems employing pigmented elastomers or embedded surface markers \cite{zhang2022hardware, rayamane2022design}.

The present work demonstrates that durability need not be a limiting factor for vision-based tactile sensing in industrial robotics. By combining a highly wear-resistant protective surface with a replaceable sensing cartridge and a failure mode that preserves useful tactile function well beyond the onset of damage, the proposed design addresses several practical barriers that have limited wider deployment of vision-based tactile sensors in harsh operating environments. 

\FloatBarrier
\section{Conclusion}
\label{sec:conclusion}
Conclusions
Vision-based tactile sensing has demonstrated significant potential for robotic manipulation, but limited durability remains a barrier to its use in demanding industrial environments. This work investigated whether a practical vision-based tactile sensor could combine increased laboratory durability, gradual failure behavior, and rapid field replacement.

The developed sensor reached the protective-film rupture endpoint after approximately 2–3 hours in the accelerated sanding test. During repetitive probe testing, all nine sensors remained functionally usable when testing was discontinued: seven after 5 days, one after 6 days, and one after 8 days. Under the same laboratory protocols, the commercial comparison specimens exhibited initial surface-film rupture more than two orders of magnitude earlier. Because functional failure of the developed sensors was not observed, the repetitive-loading results represent lower bounds on durability under the defined test conditions.

In addition to increased durability, the sensor exhibited gradual rather than catastrophic damage. Abrasion produced localized rupture of the TPU protective film while much of the sensing surface remained functional. Repetitive loading produced circular probe imprints and, in some specimens, small localized delamination blisters, but useful tactile imaging was preserved through the respective test endpoints. This behavior could provide warning of developing damage and allow maintenance to be scheduled before complete loss of sensing capability.

The sensing surface was designed as a replaceable cartridge that can be exchanged without sensor disassembly, electrical disconnection, optical realignment, or recalibration. Removal requires only a simple plastic prying tool. Combined with gradual failure behavior, this architecture offers the potential for scheduled maintenance and economical replacement, both of which are important for industrial robotic deployment.

Although demonstrated using a nonpigmented vision-based tactile sensor, the concepts investigated here—including a durable protective surface, gradual degradation, a replaceable sensing cartridge, and accelerated comparative durability testing—may also be applicable to other vision-based tactile sensor architectures.

Future work will focus on increasing resistance to interfacial delamination during repetitive loading, improving abrasion-test repeatability through the use of standardized wear media, and investigating whether graded TPU structures can reduce interfacial stress while preserving tactile sensitivity.

Rather than attempting to eliminate wear entirely, this work demonstrates an approach in which wear is reduced, damage develops gradually, and the worn sensing surface can be replaced rapidly. The results suggest that maintainable sensor construction offers a practical path toward vision-based tactile sensing for demanding robotic applications.



\acknowledgments{
Paul Souza of Accurate Tool, Lakeville, Massachusetts, fabricated the benchtop injection-molding machine and the molds used to manufacture the cartridges for both tactile fingers shown in \autoref{fig:humanoid_vbts}. He also fabricated the silicone-gel casting molds shown in \autoref{fig:manufacturing_components}, which were used to produce all VBTS specimens evaluated in this study.
}


\bibliography{example}  

\end{document}